\documentclass[lettersize,journal]{IEEEtran}
\usepackage{amsmath,amsfonts}
\usepackage{algorithmic}
\usepackage{algorithm}
\usepackage{array}
\usepackage[caption=false,font=normalsize,labelfont=sf,textfont=sf]{subfig}
\usepackage{textcomp}
\usepackage{stfloats}
\usepackage{url}
\usepackage{verbatim}
\usepackage{graphicx}
\usepackage{cite}
\usepackage[caption=false]{subfig}
\usepackage[table]{xcolor}
\usepackage{stfloats}

\usepackage{amssymb}
\usepackage{float}
\usepackage{array}
\usepackage{booktabs}
\usepackage{bbding}
\usepackage{booktabs}

\usepackage{lineno}
\usepackage{multirow}
\usepackage{amsmath,amssymb,amsfonts}
\usepackage{algorithmic}
\usepackage{graphicx}
\usepackage{textcomp}
\usepackage{color}
\usepackage{hyperref}

\begin{document}
	
	\title{SpeGCL: Self-supervised Graph Spectrum Contrastive Learning without Positive Samples}
	
	
\author{Yuntao~Shou,
	Xiangyong Cao, and Deyu Meng
	\\
	\thanks{Corresponding Author: Xiangyong Cao (caoxiangyong@mail.xjtu.edu.cn)}
	\IEEEcompsocitemizethanks{\IEEEcompsocthanksitem Y. Shou, and X. Cao are with School of Computer Science and Technology, Xi’an Jiaotong University, Xi’an, China, Ministry of Education Key Laboratory for Intelligent Networks and Network Security, Xi’an Jiaotong.
		(shouyuntao@stu.xjtu.edu.cn,
		caoxiangyong@mail.xjtu.edu.cn,
		). D. Meng is with School of Mathematics and Statistics, Xi’an Jiaotong University, Xi’an, China. (dymeng@mail.xjtu.edu.cn)}
}
	
	\maketitle
	
	\begin{abstract}
		Graph Contrastive Learning (GCL) excels at managing noise and fluctuations in input data, making it popular in various fields (e.g., social networks, and knowledge graphs). Our study finds that the difference in high-frequency information between augmented graphs is greater than that in low-frequency information. However, most existing GCL methods focus mainly on the time domain (low-frequency information) for node feature representations and cannot make good use of high-frequency information to speed up model convergence. Furthermore, existing GCL paradigms optimize graph embedding representations by pulling the distance between positive sample pairs closer and pushing the distance between positive and negative sample pairs farther away, but our theoretical analysis shows that graph contrastive learning benefits from pushing negative pairs farther away rather than pulling positive pairs closer. To solve the above-mentioned problems, we propose a novel spectral GCL framework without positive samples, named SpeGCL. Specifically, to solve the problem that existing GCL methods cannot utilize high-frequency information, SpeGCL uses a Fourier transform to extract high-frequency and low-frequency information of node features, and constructs a contrastive learning mechanism in a Fourier space to obtain better node feature representation. Furthermore, SpeGCL relies entirely on negative samples to refine the graph embedding. We also provide a theoretical justification for the efficacy of using only negative samples in SpeGCL. Extensive experiments on un-supervised learning, transfer learning, and semi-supervised learning have validated the superiority of our SpeGCL framework over the state-of-the-art GCL methods.
	\end{abstract}
	
	
	\section{Introduction}
	
	\IEEEPARstart{T}{he} proliferation of social networks and the advent of vast graph datasets have propelled Graph Neural Networks (GNNs) to the forefront as a potent tool for graph data processing and knowledge extraction. GNNs are now extensively utilized in various sectors \cite{10502132, 10505804, shou2022conversational, shou2025masked, shou2023comprehensive, meng2023deep, meng2024multi, meng2024deep}, including recommendation systems \cite{deldjoo2023review}, bioinformatics \cite{li2022geomgcl}, and a myriad of other domains \cite{wang2023correntropy, shou2023low, shou2024adversarial, ai2024gcn, ai2023gcn, ai2023two}. Traditionally, GNNs have been optimized through supervised learning, which is heavily dependent on high-quality, expert-annotated labels. However, acquiring such detailed labels necessitates significant domain expertise and is resource-intensive. To address these challenges, approaches such as Variational Graph Autoencoder (VGAE) \cite{kipf2016variational} and Graph Sample and Aggregation (GraphSAGE) \cite{hamilton2017inductive} have been developed to facilitate unsupervised learning by reconstructing the adjacency matrix of the graph. Additionally, the DeepWalk \cite{perozzi2014deepwalk} algorithm employs a random walk strategy to generate node embedding representations in a self-supervised manner, further enhancing the capabilities of GNNs without the need for extensive manual labelling.
	
	Recently, with the development of graph contrastive learning (GCL), the performance of some self-supervised training methods is comparable to supervised learning methods~\cite{liu2022revisiting, yin2022autogcl}. Specifically, GCL operates by creating various graph perspectives through data augmentation, an approach that minimizes the distance between input positive pairs in feature space and maximizes the distance between negative pairs. For instance, Deep Graph Infomax (DGI) \cite{velivckovic2018deep} leverages mutual information (MI) to enhance the model's ability to distill valuable insights from the node's local context. Meanwhile, Graph Contrastive Learning (GraphCL) \cite{you2020graph} aims to refine node representations so that they more accurately reflect the graph's structural and semantic attributes within the embedding space through contrastive techniques. Additionally, Spectral Feature Augmentation (SFA) \cite{zhang2023spectral} employs feature-level augmentation to estimate low-rank feature approximations across different graphs, offering a complementary strategy to other existing graph augmentation methods.
	
	\begin{figure}
		\centering
		\includegraphics[width=0.5\textwidth]{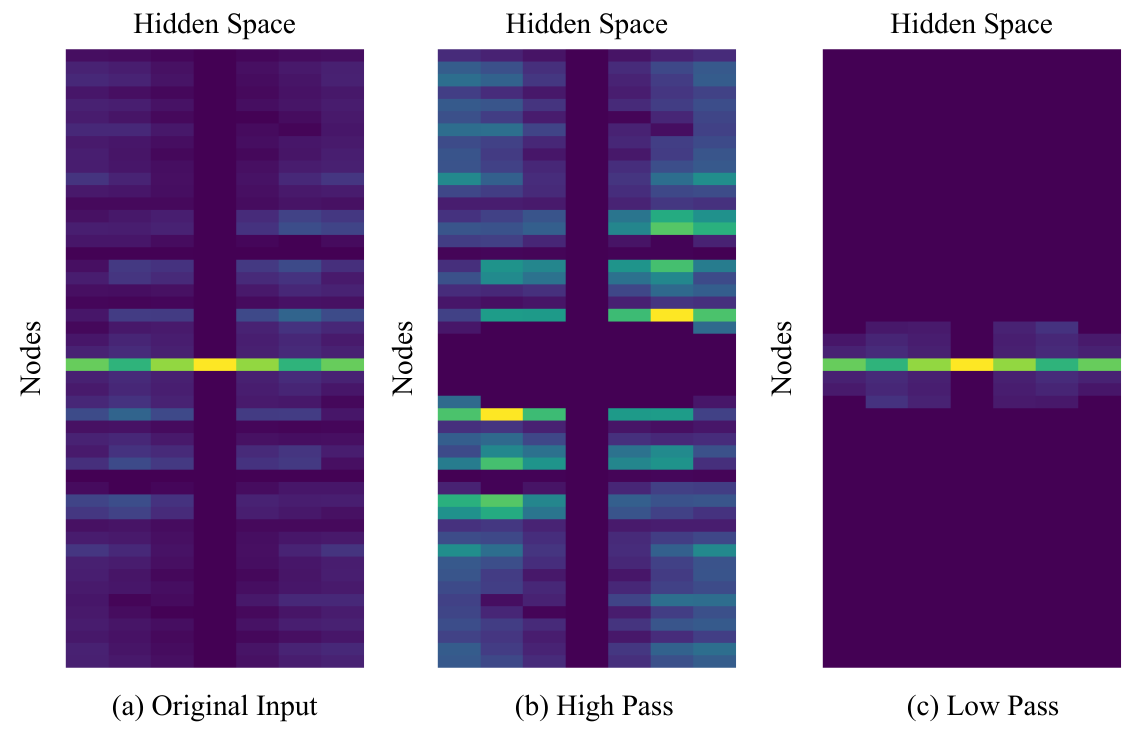}
		\caption{Visualization of the original features, low-frequency features, and high-frequency features on the MUTAG dataset in the frequency domain.}
		\label{fig:freq}
		\vspace{-4mm}
	\end{figure}
	
	As depicted in Figure \ref{fig:freq}, we notice that the low-frequency components exhibit relatively mild variations, whereas the high-frequency components undergo significant changes. This observation leads us to posit that high-frequency components are pivotal in GCLs, given the substantial disparities between each "pixel". SpCo \cite{liu2022revisiting} also has theoretically established that high-frequency information holds greater significance than low-frequency information in GCL. Nonetheless, SpCo necessitates eigendecomposition of the Laplacian matrix, leading to considerable computational overhead (i.e., $O(n^3)$). However, many current GCL methods focus on feature transformation in the time domain and fail to capture the high-frequency aspects of node features. Furthermore, existing GCLs methods mainly obtain better node feature representation by sampling positive and negative samples pairs, but our theoretical analysis shows that graph contrastive learning actually benefits from pushing negative pairs farther away rather than pulling positive pairs closer. Drawing inspiration from SpCo, we propose a novel spectral graph contrastive learning framework, named SpeGCL, to address the aforementioned issues. In our approach, we regard the embedded representations of historical interactions between nodes as self-supervised signals and utilize Fourier transform \cite{frigo1998fftw} to isolate both low-frequency and high-frequency components of node embeddings. Furthermore, we construct multiple graph contrastive views to preserve the most expressive information within node embeddings. Contrary to prior GCL methods \cite{mo2022simple} that concurrently sample both positive and negative pairs for contrastive learning, we contend that the contrastive learning mechanism primarily relies on negative sample pairs for parameter tuning. We have also provided a theoretical demonstration that the model can achieve convergence utilizing solely negative samples.
	
	Our contributions can be summarized as follows.
	
	\begin{itemize}
		\item We propose a novel spectral graph contrastive learning (SpeGCL) model that leverages Fourier operations to concurrently harness the low-frequency and high-frequency information of nodes. Additionally, the model employs the convolution theorem to facilitate the aggregation of node features. This approach enhances the representation ability of the nodes.
		
		\item We propose a new contrastive learning strategy to train graph views, which uses only negative samples to accelerate model training and parameter optimization. We also proved that the model can converge using only negative samples.
		
		\item We extensively evaluate the proposed method SpeGCL on multiple graph classification settings. Experimental results demonstrate the superiority of SpeGCL compared with other state-of-the-art GCL methods.
		
	\end{itemize}
	
	\section{Related work}
	
	Inspired by the remarkable success of contrastive learning in computer vision (CV) and natural language processing (NLP) \cite{chen2020simple, grill2020bootstrap, chen2021exploring, shou2023graph, shou2023czl, shou2024contrastive, shou2024revisiting, meng2024masked}, many graph contrastive learning methods (GCLs) \cite{xia2022progcl, luo2022automated, ghose2023spectral, suresh2021adversarial, zhu2021graph, shou2024efficient, meng2024revisiting, shou2023graphunet} have been proposed in recent years. These methods introduce data augmentation strategies, utilize the perturbations of nodes and edges in the graph structure, generate two augmented views, and learn graph representations by maximizing the mutual information (MI) between the two views. Specifically, the core idea of GCLs is to capture the structural information and semantic features in the graph by comparing different graph views, thereby improving the representation ability of the model. For example, Deep Graph Infomax (DGI) \cite{velivckovic2018deep}, as one of the early representative methods, adopts the InfoMax loss function to improve the graph representation learning effect by maximizing the mutual information between the representation of the correct node in the graph and the representation of other nodes. DGI emphasizes the learning of global graph representations, aiming to improve the model's understanding of the entire graph structure. Different from DGI, InfoGraph \cite{sun2019infograph} focuses on comparing the graph representations of different substructures, which can not only capture the characteristics of the global graph structure, but also obtain the fine-grained information of local nodes, thereby optimizing the representation learning of nodes and substructures at different levels. GCC \cite{qiu2020gcc} learns a common representation that can be generalized in multiple graphs by designing cross-graph comparison tasks. This method adopts a structure-based graph data augmentation strategy and improves generalization ability by maximizing local and global information between nodes. Sub-GCL \cite{jiao2020sub} enhances graph representation by learning comparisons between subgraphs. This method proposes to extract subgraphs from the global graph and designs subgraph comparison tasks to capture different levels of graph information. Sub-GCL improves the sensitivity of graph models to local structures by comparing the representations of different subgraphs. InfoGCL \cite{xu2021infogcl} proposes a graph comparison learning framework based on information theory. The key to this method is to automatically select important graph structure features to participate in comparison learning through a learnable selection mechanism. By introducing different graph enhancement strategies, InfoGCL can adaptively select the structural information that best represents the graph, thereby better capturing the key information in the graph. MVGRL \cite{hassani2020contrastive} is a multi-view graph comparison learning method that generates multiple views and performs comparison learning between different views to improve the robustness of graph representation.
	
	Another influential model is GRACE \cite{zhu2020deep}, whose core idea is to improve representation capabilities at the node level by maximizing the similarity between positive contrast terms and minimizing the similarity between negative contrast terms. Similarly, GraphCL \cite{you2020graph} focuses on learning graph-level representations, and by maximizing the mutual information between different enhanced views, the graph model can capture the global structural characteristics of the graph.
	
	Based on these pioneering works, new GCL methods have been proposed in recent years, which have made significant progress in learning both graph-level representations and node-level representations. However, unlike the above methods, our work is not limited to designing specific graph enhancement views. Instead, we explore whether it is necessary to rely on high-frequency information in the process of graph representation learning from a broader graph spectrum perspective. We try to reveal the role of high-frequency information in graph contrastive learning and propose a new framework that enables the model to more effectively utilize different frequency information in the graph, thereby improving the quality of representation learning.
	
	\textbf{Frequency-domain Deep Learning.} The frequency domain analysis method has always been a classic tool in the field of traditional signal processing \cite{reddy2018digital, pitas2000digital}. Through frequency domain techniques such as Fourier transform, signals can be converted into frequency domain space, so that the frequency components and structural characteristics of the signal can be better understood. In traditional signal processing, frequency domain analysis is widely used in audio processing, image processing, communication systems and other fields. Recently, with the development of deep learning technology, frequency domain methods have begun to be used to analyze the optimization \cite{xu2019training, yin2019fourier} and generalization capabilities \cite{wang2020high, xu2018understanding} of deep neural networks. The successful application of frequency domain methods in the field of deep learning may be because the input of DNNs can be regarded as signal data, and the training process of the model can be regarded as a signal processing process \cite{xu2020frequency}. In addition to analyzing the optimization and generalization capabilities of DNNs, frequency domain methods are also integrated into DNNs to learn non-local \cite{chi2020fast, li2020fourier} or domain-generalizable representations \cite{lin2023deep}. This integrated approach allows deep learning models to extract more global feature information from the data and have better generalization capabilities.
	
	\section{Preliminarties}
	
	\subsection{Notations}
	We assume that a graph is represented as $\mathcal{G}=\{\mathcal{V},\mathcal{E}\}$, where $\mathcal{V}=\{v_1,v_2,\ldots,v_N\}$ represents the set of nodes and $\mathcal{E} \in \mathcal{V} \times \mathcal{V}$ represents the set of edges. $X=\{x_i\}^N_{i=1}$ and $A \in \{0,1\}^{N \times N}$ are the feature matrix and adjacency matrix of the graph, where $x_i$ represents the feature vectors of node $v$, $a_{ij}=1$ indicates that there is an edge relationship between $v_i$ and $v_j$, otherwise $a_{ij}=0$.
	
	\subsection{Fourier Transform}
	Fourier transform \cite{soliman1990continuous} is widely used in signal processing, which can convert time domain signals into frequency domain signals. In this article, we use Discrete Fourier Transform (DFT) to perform signal conversion as follows:
	\begin{equation}
		\mathcal{F}(m,n)=\sum_{x=0}^{M-1}\sum_{y=0}^{N-1}f(x,y)e^{-j2\pi\left(\frac{m}Mx+\frac{n}Ny\right)}
	\end{equation}
	where $j$ represents the imaginary unit, $f(x,y)$ represents the time domain signals, $\mathcal{F}(m,n)$ represents the frequency domain signals, $(m,n)$, and $(x,y)$ is the coordinates of the Fourier space and time domain space, respectively. $\mathcal{F}^{-1}(x)$ is the inverse Fourier transform. We reconstruct the original signals via the IDFT:
	\begin{equation}
		f(x,y)=\frac1{MN}\sum_{m=0}^{M-1}\sum_{z=0}^{N-1}\mathcal{F}(m,n)e^{j2\pi\left(\frac{m}Mx+\frac{n}Ny\right)}
	\end{equation}
	
	Since the computational complexity of DFT/IDFT is large and difficult to adapt to large-scale data sets, in this paper we apply fast Fourier transform (FFT) and inverse fast Fourier transform (IFFT) to reduce the complexity from $O(n^2)$ to $O(nlogn)$. The amplitude component $\mathcal{A}(m,n)$ and phase
	component $\mathcal{P}(m,n)$ is defined as follows:
	\begin{equation}
		\begin{aligned}
			&\mathcal{A}(m,n)=\mathcal{R}^2(m,n)+\mathcal{I}^2(m,n)\\
			&\mathcal{P}(m,n)=\arctan[\frac{\mathcal{I}(m,n)}{\mathcal{R}(m,n)}]
		\end{aligned}
	\end{equation}
	where $\mathcal{R}^2(m,n)$ and $\mathcal{I}^2(m,n)$ are the real and
	imaginary parts respectively.
	
	\subsection{Training Objective}
	
	The main goal of GCLs \cite{velivckovic2018deep, hassani2020contrastive, zhang2021canonical} is to learn discriminative embeddings without supervision. The method is to generate two augmented views in a predefined way (e.g., masked nodes and edge perturbations, etc.) and encode them by GCN to obtain the node embeddings of the two augmented views. Subsequently, for a target node, its embedding in an enhanced view is designed to be close to its positive samples and far away from its negative samples. The GCLs method \cite{zhu2021graph, you2020graph} uses the classic InfoNCE loss \cite{gutmann2010noise} as the optimization objective to distinguish similar nodes from dissimilar nodes. The optimization objective is defined as follows:
	\begin{equation}\begin{aligned}
			\mathcal{L}_{\mathrm{NCE}}& \triangleq-\log\frac{e^{f(x)^Tf(y)/\tau}}{e^{f(x)^\mathsf{T}f(y)/\tau}+\sum_ie^{f(x)^\mathsf{T}f(y_i^-)/\tau}}  \\
			&=\underbrace{-\frac1\tau f(x)^{\mathsf{T}}f(y)}_{\mathrm{alignment}} 
			+\underbrace{\log(e^{f(x)^{\mathsf{T}}f(y)/\tau}+\sum_ie^{f(x)^{\mathsf{T}}f(y_i^{-})/\tau})}_{\text{uniformity}}
	\end{aligned}\end{equation}
	where $(x, y) \sim p_{pos}$ is the positive pair, $p_{pos}$ is the probability distribution of the positive pair, $\tau$ is a decay coefficient, and $\{y_i^-\}_{i=1}^M\overset{\mathrm{i.i.d.}}{\operatorname*{\sim}}p_\mathrm{y}$ is the negative samples.
	
	\section{Proposed Method}
\begin{figure*}
		\centering
		\includegraphics[width=1\linewidth]{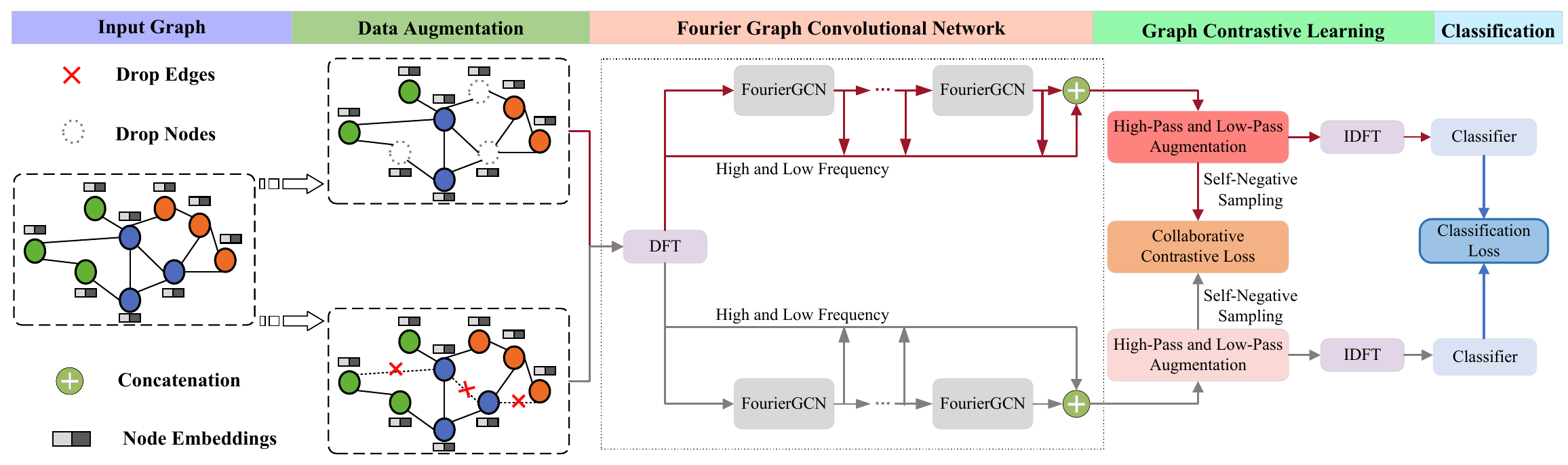}
		\caption{The overall architecture of the proposed SpeGCL model. Specifically, we use node masking and edge perturbation strategies for data augmentation and use DFT to separate the high-pass and low-pass components of the augmented view. Then we aggregate the node information based on the convolution theorem and only sample negative samples to construct the contrastive loss without positive samples. In particular, we use IDFT for time-frequency transformation in a semi-supervised experimental setting.}
		\label{fig:archicture}
		\vspace{-4mm}
	\end{figure*}

	As shown in Fig. \ref{fig:archicture}, the overall process of the proposed SpeGCL method includes four modules: data augmentation, Fourier graph convolutional neural network, contrastive learning and graph classification. In the following sections, we will describe their implementation process in detail.
	
	\subsection{Generated Multi-view Augmentation}

	\textbf{Node-Masking View}
	We perform automatic learnable node masking before each information aggregation and feature update of GCN to generate the augmented node-masking views. The node-masking view is as:
	\begin{equation}
		\mathcal{G}_{N D}^{(l)}=\left\{\left\{v_i \odot \eta_i^{(l)} \mid v_i \in \mathcal{V}, \mathcal{E}\right\}\right\}
	\end{equation}
	where $\eta_i^{(l)} \in \{0,1\}$ is sampled from a parameterized Bernoulli distribution $Bern(\omega_i^l)$, and $\eta_i^{(l)} = 0$ represents masking node $v_i$, $\eta_i^{(l)} = 1$ represents keeping node $v_i$.
	
	\textbf{Edge Perturbation View}
	Edge perturbation can be seen as a subtle adjustment to the original graph structure to create a new graph view that enables the model to better understand the relationship between nodes during training and improve its robustness. By properly perturbing the edges, useful edge information is retained and redundant or erroneous connections are removed to enhance the graph structure's ability to express downstream tasks. Specifically, the generation of edge perturbation views can be described by the following formula:
	\begin{equation}
		\mathcal{G}_{E D}^{(l)}=\left\{\mathcal{V},\left\{e_{i j} \odot \eta_{i j}^{(l)} \mid e_{i j} \in \mathcal{E}\right\}\right\}
	\end{equation}
	where $\eta_
	{ij}^{(l)} \in \{0,1\}$ is also sampled from a parameterized Bernoulli distribution $Bern(\omega_{ij}^l)$, and $\eta_{ij}^{(l)} = 0$ represents perturbating edges $e_{ij}$, $\eta_i^{(l)} = 1$ represents keeping edge $e_{ij}$.
	
	\subsection{Fourier Graph Convolutional Network}
	The classical graph networks (e.g., GCN \cite{kipfsemi} and GAT \cite{velivckovicgraph}) cannot compute in the frequency domain and obtain feature representations of hidden layer nodes, and their computational complexity is high (quadratic complexity). Therefore, we design a more efficient and effective method to obtain the feature representation of nodes within Fourier space based on the convolution theorem \cite{katznelson2004introduction}. The graph convolution operation can be rewritten as follows:
	\begin{equation}\begin{aligned}
			\mathcal{F}\left(X\right)\mathcal{F}\left(\kappa\right)& =\mathcal{F}\left(\left(X*\kappa\right)[i]\right) 
			\\&=\mathcal{F}\left(X\left[j\right]\kappa\left[i-j\right]\right)=\mathcal{F}\left(X\left[j\right]\kappa\left[i,j\right]\right) \\
			&=\mathcal{F}\left(A_{ij}X\left[j\right]W\right)=\mathcal{F}\left(AXW\right)
	\end{aligned}\end{equation}
	where $\left(X*\kappa\right)[i]$ represents the convolution of $X$ and $\kappa$ in the fourier spaces, and $\kappa[i,j] = A_{ij} W$.
	
	\subsection{Graph Contrastive Learning}
	The low-frequency bias of deep learning models limits the usefulness of graph encoders \cite{xu2019frequency}. To solve the above problems, we constructed samples containing low-frequency information and high-frequency information for graph contrastive learning to improve the feature discrimination ability of the encoder. Unlike previous GCL work \cite{mo2022simple} that used positive and negative pairs to achieve contrastive learning, we only use negative pairs.
	
	\subsubsection{Data Augmentation Operators}
	We design high-pass enhancement and low-pass augmentation to obtain high-frequency and low-frequency features of node features. Specifically, we first calculate the frequency domain representation $X^{Freq} \in \mathbb{R}^{N \times d}$ of the nodes features $X \in \mathbb{R}^{N \times N}$ as follows:
	
	\begin{equation}
		X^{Freq}=\mathrm{FShift}(\mathcal{F}(X))
	\end{equation}
	where $\mathcal{F}(\cdot)$ is the fast Fourier transform, and $\mathrm{FShift}(\cdot)$ indicates that the zero-frequency component of the converted frequency domain moves toward the center $(\frac{N}{2}, \frac{d}{2})$.
	
	\textbf{Low-Pass Augmentation (LPA).} LPA adopts the low-frequency component as the feature representation of the node and it is close to the central part of Xfreq. Therefore, we set the threshold of the low-frequency component to obtain the low-pass component, which can be formalized as follows:
	
	\begin{equation}
		X_{aug}^{freq}=\mathrm{LPA}(m,z)\cdot X^{freq}
	\end{equation}
	and 
	\begin{equation}
		\mathrm{LPA}(m,z)=\left\{\begin{array}{c}1,D(m,z)\leq D^L\\0,D(m,z)>D^L\end{array}\right.
	\end{equation}
	where $(m,z)$ is the coordinate position in the frequency domain,  $D_L$ is the low-frequency threshold, $D(m, z)$ represents the distance between point $(m, z)$ and the center point $(\frac{N}{2}, \frac{d}{2})$, which can be formalized as follows:
	\begin{equation}
		D(m,z)=\sqrt{(m-\frac{N}{2})^2+(z-\frac{d}{2})^2}
	\end{equation}
	
	LPA only retains the areas where the signal changes gently in the node features while filtering out the noise in the features.
	
	\textbf{High-Pass Augmentation (HPA).} HPA retains high-frequency information in node features, representing rapidly changing areas. We regard the part of the point $(m, z)$ greater than the low-frequency threshold $D^L$ as high-frequency information. We perform contrastive learning on the constructed high-frequency and low-frequency components to alleviate the low-frequency preference problem mentioned in the F principle \cite{xu2019frequency}. In addition, the encoder can filter the noisy information of the nodes.
	
	\subsubsection{Self-Negative Sampling}
	In this section, we propose a self-supervised GCL framework without positive samples. Specifically, we first analyze the traditional NCE loss. Then, we further derive the self-supervised NCE loss for self-negative sampling.
	
	Previous research \cite{wang2020understanding} found that the NCE loss function has some important asymptotic properties as follows:
	
	\textbf{Theorem 1.} For a fixed $\tau > 0$ , when the number of negative samples $M \rightarrow \infty$, the contrastive loss $\mathcal{L}_{NCE}$ converges and the absolute deviation decays with $O(M^{-2/3}$). If there exists a perfectly uniform encoder $f$, it can obtain the minimum value \cite{wang2020understanding}. 
	
	According to our point of view, the aligned parts should be semantically similar, i.e., $f(x)^Tf(y) \rightarrow 1$. Therefore, we believe that the main task of contrastive learning is to optimize the uniformity part in Eq. 6. We can improve NCE Loss and get the theoretical bound.
	
	\textbf{Proposition 1.} For fixed $\tau > 0$ , the upper limit of $\mathcal{L}_U$ is always controlled by $\mathcal{L}_{NCE}$:
	\begin{equation}
		\begin{aligned}
			\mathcal{L}_U& =-\frac1\tau+\underset{{\{y_i^-\}_{\boldsymbol{i}=1}^M\overset{\mathrm{i.i.d.}}{\operatorname*{\sim}}\boldsymbol{p}\boldsymbol{y}}}{\mathbb{E}}\left[\log(e^{1/\tau}+\sum_ie^{f(x)^{\mathsf{T}}f(y_i^-)/\tau})\right]  
			\\&\leq\mathcal{L}_{\mathrm{NCE}}.
		\end{aligned}
	\end{equation}
	
	By optimizing $\mathcal{L}_U$, the model can achieve the effect of pushing dissimilar nodes farther away and similar nodes relatively close. In other words, even if we cannot draw similar nodes closer, we can ensure that the model pushes those dissimilar nodes far enough away. Therefore, we only focus on pushing negative samples further apart. 
	
	Based on the above analysis, we prove that the focus of GCL is to sample negative sample pairs. We find that a large number of negative samples is crucial for the convergence of GCL. The improved NCE loss function has important asymptotic properties as follows:
	
	\textbf{Theorem 2.} For the given constant $\tau\in\mathbb{R}^+$, $L_U$ still converges to the same limit as NCE loss, and the absolute deviation decays by $O(M^{-2/3})$.
	
	
	\section{Experiments}
	\label{sec:Experiments}
	
	\subsection{Datasets} We use the TUDataset dataset\footnote{https://chrsmrrs.github.io/datasets/docs/datasets/} \cite{morris2020tudataset} to verify the effectiveness of the proposed SpeGCL under experimental settings of unsupervised and semi-supervised learning. Under the experimental setting of transfer learning, we pre-trained on the ChEMBL dataset \cite{mayr2018large} and fine-tuned the model using the MoleculeNet dataset\footnote{http://snap.stanford.edu/gnn-pretrain/} \cite{wu2018moleculenet}. The detail information of those used datasets can be found in Tables \ref{tab:1}, and \ref{tab:2}.
	
	\begin{table}[htbp]
		\centering
		\caption{Statistics of TU-datasets and OGB dataset.}
		\label{tab:1}
		\setlength{\tabcolsep}{6pt}{
			\begin{tabular}{lcccc}
				\toprule
				\textbf{Dataset} & Graphs & Avg Nodes & Avg Edges & Class        \\ \midrule
				MUTAG            & 188      & 17.93     & 19.79     & 2         \\
				PROTEINS         & 1,113    & 39.06     & 72.82     & 2         \\
				NCI1             & 4,110    & 29.87     & 32.3      & 2         \\
				DD               & 1,178    & 284.32    & 715.66    & 2         \\
				COLLAB           & 5,000    & 74.49     & 2457.78   & 3      \\
				IMDB-B           & 1,000    & 19.77     & 96.53     & 2      \\
				REDDIT-B            & 2,000    & 429.63    & 497.75    & 2      \\
				REDDIT-M-5K      & 5000   & 508.5      & 492      & 2        \\ \bottomrule
		\end{tabular}}
	\end{table}
	
	\begin{table}[htbp]
		\centering
		\caption{Statistics of MoleculeNet datasets.}
		\label{tab:2}
		\setlength{\tabcolsep}{7pt}{
			\begin{tabular}{lcccc}
				\toprule
				\textbf{Model} & Graphs    & Avg Nodes & Avg Degree & \#Tasks     \\ \midrule
				BBBP           & 2,039     & 24.06     & 51.9       & 1        \\
				Tox21          & 7,813     & 18.57     & 38.58      & 12       \\
				ToxCast        & 8,576     & 18.78     & 38.62      & 617      \\
				SIDER          & 1,427     & 33.64     & 70.71      & 27       \\
				ClinTox        & 1,477     & 26.15     & 55.76      & 2        \\
				MUV            & 93,087    & 24.23     & 52.55      & 17       \\
				HIV            & 41,127    & 25.51     & 54.93      & 1        \\
				BACE           & 1,513     & 34.08     & 73.71      & 1        \\ \bottomrule
		\end{tabular}}
	\end{table}
	
	\subsection{Evaluation Protocols} To evaluate the effectiveness of the proposed SpeGCL, we conduct extensive experiments under different experimental settings and different datasets. Specifically, for unsupervised learning and semi-supervised learning tasks, we selected multiple datasets in TUDataset \cite{morris2020tudataset}, which cover graph data of various social networks and biochemical molecules. We report the mean test accuracy by 10-fold cross-validation with the standard deviation as the final performance. In terms of transfer learning, we first performed pre-training on the ChEMBL dataset \cite{mayr2018large}, which is a graph dataset containing a large amount of biological activity information. Next, we use the MoleculeNet dataset \cite{wu2018moleculenet} to fine-tune the model.
	
	\subsection{Baselines} Under the unsupervised learning setting, we compare the proposed SpeGCL with the kernel-based methods like GL \cite{shervashidze2009efficient},
	WL \cite{shervashidze2011weisfeiler}, and DGK \cite{yanardag2015deep}, and the graph representation methods like node2vec \cite{grover2016node2vec}, sub2vec \cite{adhikari2018sub2vec}, and graph2vec \cite{narayanan2017graph2vec}, and the graph contrastive methods like InfoGraph \cite{sun2019infograph}, GraphCL \cite{you2020graph}, JOAOv2 \cite{you2021graph}, AD-GCL \cite{suresh2021adversarial}, AutoGCL \cite{yin2022autogcl}, SEGA \cite{wu2023sega}, GCS \cite{wei2023boosting}, and LAMP-Soft \cite{chen2024uncovering}. Under the transfer learning setting, we compare the proposed SpeGCL with the graph contrastive methods like Infomax \cite{velivckovic2018deep}, EdgePred \cite{hu2020strategies}, AttrMasking \cite{hu2020strategies}, ContextPred \cite{hu2020strategies}, GraphCL \cite{you2020graph}, JOAOv2 \cite{you2021graph}, AD-GCL \cite{suresh2021adversarial}, SEGA \cite{wu2023sega}, GCS \cite{wei2023boosting}, and LAMP-Soft \cite{chen2024uncovering}. Under the semi-supervised learning setting, we compare the proposed SpeGCL with the graph contrastive methods like GCA \cite{zhu2021graph}, GraphCL \cite{you2020graph}, JOAOv2 \cite{you2021graph}, and AD-GCL \cite{suresh2021adversarial}, and SEGA \cite{wu2023sega}.
	
	\textbf{GL.} GL \cite{shervashidze2009efficient} is a feature extraction and similarity measurement method for graph-structured data analysis. It describes the overall structure and characteristics of a graph based on small topological patterns in the graph.
	
	\textbf{WL.} WL\footnote{https://github.com/BorgwardtLab/WWL?tab=readme-ov-file} \cite{shervashidze2011weisfeiler} is a kernel method for graph data analysis, which calculates the similarity between graphs through layer-by-layer iterative labeling and aggregation of the graph's structure.
	
	\textbf{DGK.} DGK\footnote{https://github.com/pankajk/Deep-Graph-Kernels} \cite{yanardag2015deep} combines the advantages of deep learning and graph kernel methods, and can retain the structural information.
	
	\textbf{Node2vec.} The basic idea of Node2Vec\footnote{https://github.com/eliorc/node2vec} \cite{grover2016node2vec} is to learn the vector representation of nodes by performing random walks on the graph and then using the Word2Vec model.
	
	\textbf{Sub2vec.} The basic idea of the Sub2Vec\footnote{https://github.com/bijayaVT/sub2vec} algorithm is to treat substructures in the graph (e.g., subgraphs, subtrees, etc.) as words, and then learn the vector representation of the substructure through the context of the substructure (e.g.,  adjacent nodes, edges, etc.).
	
	\textbf{Graph2vec.} Graph2vec\footnote{https://github.com/benedekrozemberczki/graph2vec} \cite{narayanan2017graph2vec} is a method for learning graph embeddings that is able to map the entire graph into a low-dimensional vector space and preserve the structural and semantic information of the graph.
	
	\textbf{InfoGraph.} The core idea of InfoGraph\footnote{https://github.com/sunfanyunn/InfoGraph} \cite{sun2019infograph} is to encode the transformed graph by using a GNN model and compare the encoding result with the original graph to maximize the similarity between them.
	
	\textbf{GCA.} GCA\footnote{https://github.com/CRIPAC-DIG/GCA} uses adaptive graph data augmentation to generate different views of the graph for contrastive learning. Traditional GCLs usually relie on pre-set random augmentation strategies, while GCA dynamically adjusts these augmentation operations based on the structural characteristics and the importance of the nodes, allowing the model to learn more effective graph representations from more relevant perspectives.
	
	\textbf{GraphCL.} GraphCL\footnote{https://github.com/Shen-Lab/GraphCL} \cite{you2020graph} constructs local contrastive tasks and global contrastive tasks to maximize the similarity.
	
	\textbf{JOAOv2.} The key innovation of JOAOv2\footnote{https://github.com/Shen-Lab/GraphCL\_Automated} \cite{you2021graph} is the introduction of a connection-based graph embedding optimization framework, in which the connection relationships between nodes are treated as a hypersphere in the embedding space. The optimization goal is to maximize the overlapping area of the connection areas between adjacent nodes and to minimize the overlapping area of the connection areas between non-adjacent nodes.
	
	\begin{table*}[htbp]
		\centering
		\caption{Overall comparison with existing unsupervised learning methods on multiple graph classification datasets. We report accuracy results as mean $\pm$ std.}
		\label{tab:unsuperise}
		\renewcommand\arraystretch{1}
		\setlength{\tabcolsep}{2.5mm}{
			\begin{tabular}{lcccccccc}
				\toprule
				Model     & MUTAG       & PROTEINS   & DD         & NCI1       & COLLAB     & IMDB-B     & REDDIT-B   & REDDIT-M-5K \\ \midrule
				GL \cite{shervashidze2009efficient}       & 81.66±2.11  & -          & -          & -          & -          & 65.87±0.98 & 77.34±0.18 & 41.01±0.17  \\
				WL \cite{shervashidze2011weisfeiler}       & 80.72±3.00  & 72.92±0.56 & -          & 80.01±0.50 & -          & 72.30±3.44 & 68.82±0.41 & 46.06±0.21  \\
				DGK \cite{yanardag2015deep}      & 87.44±2.72  & 73.30±0.82 & -          & 80.31±0.46 & -          & 66.96±0.56 & 78.04±0.39 & 41.27±0.18  \\
				node2vec \cite{grover2016node2vec}  & 72.63±10.20 & 57.49±3.57 & -          & 54.89±1.61 & -          & -          & -          & -           \\
				sub2vec \cite{adhikari2018sub2vec}  & 61.05±15.80 & 53.03±5.55 & -          & 52.84±1.47 & -          & 55.26±1.54 & 71.48±0.41 & 36.68±0.42  \\
				graph2vec \cite{narayanan2017graph2vec} & 83.15±9.25  & 73.30±2.05 & -          & 73.22±1.81 & -          & 71.10±0.54 & 75.78±1.03 & 47.86±0.26  \\
				InfoGraph \cite{sun2019infograph} & 89.01±1.13  & 74.44±0.31 & 72.85±1.78 & 76.20±1.06 & 70.65±1.13 & 73.03±0.87 & 82.50±1.42 & 53.46±1.03  \\
				GraphCL \cite{you2020graph}  & 86.80±1.34  & 74.39±0.45 & {78.62±0.40} & 77.87±0.41 & {71.36±1.15} & 71.14±0.44 & {89.53±0.84} & 55.99±0.28  \\
				JOAOv2 \cite{you2021graph}   & -           & 71.25±0.85 & 66.91±1.75 & 72.99±0.75 & 70.40±2.21 & 71.60±0.86 & 78.35±1.38 & 45.57±2.86  \\
				AD-GCL \cite{suresh2021adversarial}    & -           & 73.59±0.65 & 74.49±0.52 & 69.67±0.51 & \textbf{73.32±0.61} & 71.57±1.01 & 85.52±0.79 & 53.00±0.82  \\
				AutoGCL \cite{yin2022autogcl}     & 88.64±1.08  & 75.80±0.36 & 77.57±0.60 & {82.00±0.29} & 70.12±0.68 & {73.30±0.40} & 88.58±1.49 & {56.75±0.18}  \\
				SEGA \cite{wu2023sega} & 90.21±0.66 & 76.01±0.42 & 78.76±0.57 & 79.00±0.72  & 74.12±0.47 & 73.58±0.44 & 90.21±0.65 &56.13±0.30 \\ 
				GCS \cite{wei2023boosting} & 90.45±0.81 & 75.02±0.39 & 77.22±0.30 &77.37±0.30  & 75.56±0.41 & 73.43±0.38 &\textbf{92.98±0.28} &57.04±0.49  \\
				LAMP-Soft \cite{chen2024uncovering} & \underline{90.89±1.04} & \underline{77.34±0.53} & \underline{80.03±0.85} &\textbf{82.17±0.48}  & \underline{75.96±0.67} & \underline{75.14±0.59} &91.63±0.55  &\underline{57.38±0.41} \\
				SpeGCL (Ours) & \textbf{91.86±2.74} & \textbf{78.05±1.23} & \textbf{81.23±0.94} & \underline{82.14±1.12} & \textbf{76.00±0.38}
				& \textbf{76.57±1.95} & \underline{91.71±0.31} & \textbf{59.44±0.18} \\
				\bottomrule
		\end{tabular}}
	\end{table*}
	
	\textbf{AD-GCL.} AD-GCL\footnote{https://github.com/susheels/adgcl} \cite{suresh2021adversarial} is a graph conrastive learning method based on adversarial graph augmentation, which uses the graph information bottleneck principle to learn graph representations that remove redundant information.
	
	\textbf{Auto-GCL.} AutoGCL\footnote{https://github.com/Somedaywilldo/AutoGCL} \cite{yin2022autogcl} is a contrastive learning method based on a learnable graph view generator that can generate more semantically similar and topologically heterogeneous comparison samples. 
	
	\textbf{SEGA.} SEGA \footnote{https://github.com/Wu-Junran/SEGA} \cite{wu2023sega} derives the definition of the anchor view, which should have the smallest structural uncertainty to ensure that the basic information of the input graph is retained.
	
	\textbf{GCS.} GCS\footnote{https://github.com/weicy15/GCS} \cite{wei2023boosting} proposes a novel self-supervised learning framework that uses gradient-based graph contrastive saliency to adaptively screen semantically relevant substructures. The most semantically discriminative structures are identified through contrastive learning, thereby generating more semantically meaningful augmented views.
	
	\textbf{LAMP-Soft.} LAMP-Soft \cite{chen2024uncovering} takes the original graph as input, dynamically generates a perturbation model by pruning the weights of the graph encoder, and performs comparative learning with the original model. In addition, in order to maintain the integrity of node embeddings, this paper designs a local contrast loss to deal with hard negative sample interference during training.
	
	\begin{table*}[htbp]
		\centering
		\caption{Overall comparison with existing transfer learning methods on multiple graph classification datasets. We report accuracy results as mean $\pm$ std.}
		\label{tab:transfer}
		\renewcommand\arraystretch{1}
		\setlength{\tabcolsep}{3mm}{
			\begin{tabular}{lcccccccc}
				\toprule
				\textbf{Model}       & BBBP       & Tox21      & ToxCast    & SIDER      & ClinTox    & MUV        & HIV        & BACE       \\ \midrule
				\textit{No Pretrain} & 65.8±4.5   & 74.0±0.8   & 63.4±0.6   & 57.3±1.6   & 58.0±4.4   & 71.8±2.5   & 75.3±1.9   & 70.1±5.4   \\ \midrule
				Infomax  \cite{velivckovic2018deep}            & 68.8±0.8   & 75.3±0.5   & 62.7±0.4   & 58.4±0.8   & 69.9±3.0   & 75.3±2.5   & 76.0±0.7   & 75.9±1.6   \\
				EdgePred \cite{hu2020strategies}            & 67.3±2.4   & 76.0±0.6   & 64.1±0.6   & 60.4±0.7   & 64.1±3.7   & 74.1±2.1   & 76.3±1.0   & 79.9±0.9   \\
				AttrMasking  \cite{hu2020strategies}        & 64.3±2.8   & 76.7±0.4   & {64.2±0.5}    & 61.0±0.7   & 71.8±4.1   & 74.7±1.4   & 77.2±1.1   & 79.3±1.6   \\
				ContextPred  \cite{hu2020strategies}        & 68.0±2.0   & 75.7±0.7   & 63.9±0.6   & 60.9±0.6   & 65.9±3.8   & 75.8±1.7   & 77.3±1.0   & 79.6±1.2   \\
				GraphCL \cite{you2020graph}             & 69.68±0.67 & 73.87±0.66 & 62.40±0.57 & 60.53±0.88 & 75.99±2.65 & 69.80±2.66 & 78.47±1.22 & 75.38±1.44 \\
				JOAOv2 \cite{you2021graph}              & 71.39±0.92 & 74.27±0.62 & 63.16±0.45 & 60.49±0.74 & {80.97±1.64} & 73.67±1.00 & 77.51±1.17 & 75.49±1.27 \\
				AD-GCL \cite{suresh2021adversarial}            & 70.01±1.07 & {76.54±0.82} & 63.07±0.72 & {63.28±0.79} & 79.78±3.52 & 72.30±1.61 & 78.28±0.97 & 78.51±0.80 \\
				AutoGCL  \cite{yin2022autogcl}              & {73.36±0.77} & 75.69±0.29 & 63.47±0.38 & 62.51±0.63 & 80.99±3.38 & 75.83±1.30 & 78.35±0.64 & 83.26±1.13 \\
				SEGA \cite{wu2023sega} & 71.86±1.06 & 76.72±0.43 & 65.23±0.91 & 63.68±0.34 & \underline{84.99±0.94} & 76.60±2.45 & 77.63±1.37 & 77.07±0.46 \\
				GCS \cite{wei2023boosting} & 71.46±0.46 & 76.16±0.41 & 65.35±0.17 & 64.20±0.35 & 82.01±1.90 & \underline{80.45±1.67} & 80.22±1.37 & 77.90±0.26 \\
				LAMP-Soft \cite{chen2024uncovering} & \underline{75.77±0.76} & \underline{77.23±0.41} & \underline{65.87±0.33} & \underline{64.24±0.68} & \textbf{85.98±1.27} & 79.50±2.19 & \textbf{81.73±1.25} & \textbf{85.58±1.43} \\
				SpeGCL (Ours)                & \textbf{76.03±0.56} & \textbf{78.31±0.18} & \textbf{66.11±0.26} & \textbf{64.73±0.42} & 84.57±2.01 & \textbf{80.61±0.97} & \underline{81.42±0.44} & \underline{84.77±1.05} \\ \bottomrule
		\end{tabular}}
	\end{table*}
	
	\begin{table*}[htbp]
		\caption{Overall comparison with existing transfer learning methods on multiple graph classification datasets. We report accuracy results as mean $\pm$ std.}
		\label{tab:tran}
		\centering
		\setlength{\tabcolsep}{6pt}{
			\begin{tabular}{l|c|ccccccccc}
				\toprule
				Pre-Train dataset          & PPI-306K    & \multicolumn{9}{c}{ZINC 2M}                                                                             \\ \midrule
				Fine-Tune dataset & PPI         & Tox21     & ToxCast   & Sider     & ClinTox   & MUV       & HIV       & BBBP      & Bace      & Average \\ \midrule
				No Pre-Train               & 64.8$\pm$1.0   & 74.6$\pm$0.4 & 61.7$\pm$0.5 & 58.2$\pm$1.7 & 58.4$\pm$6.4 & 70.7$\pm$1.8 & 75.5$\pm$0.8 & 65.7$\pm$3.3 & 72.4$\pm$3.8 & 67.1   \\
				EdgePred                   & 65.7$\pm$1.3   & \underline{76.0$\pm$0.6} & \underline{64.1$\pm$0.6} & 60.4$\pm$0.7 & 64.1$\pm$3.7 & 75.1$\pm$1.2 & \underline{76.3$\pm$1.0} & 67.3$\pm$2.4 & 77.3$\pm$3.5 & 70.1   \\
				AttrMasking                & 65.2$\pm$1.6  & 75.1$\pm$0.9 & 63.3$\pm$0.6 & \underline{60.5$\pm$0.9} & 73.5$\pm$4.3 & 75.8$\pm$1.0 & 75.3$\pm$1.5 & 65.2$\pm$1.4 & \underline{77.8$\pm$1.8} & 70.8  \\
				ContextPred                & 64.4$\pm$1.3   & 73.6$\pm$0.3 & 62.6$\pm$0.6 & 59.7$\pm$1.8 & 74.0$\pm$3.4 & 72.5$\pm$1.5 & 75.6$\pm$1.0 & \underline{70.6$\pm$1.5} & \textbf{78.8$\pm$1.2} & 70.9   \\
				GraphCL                    & \underline{67.8$\pm$0.8} & 75.1$\pm$0.7 & 63.0$\pm$0.4 & 59.8$\pm$1.3 & \textbf{77.5$\pm$3.8} & 76.4$\pm$0.4 & 75.1$\pm$0.7 & 67.8$\pm$2.4 & 74.6$\pm$2.1 & \underline{71.1}   \\
				JOAO                       & 64.4$\pm$1.3 & 74.8$\pm$0.6 & 62.8$\pm$0.7 & 60.4$\pm$1.5 & 66.6$\pm$3.1 & \underline{76.6$\pm$1.7} & \textbf{76.9$\pm$0.7} & 66.4$\pm$1.0 & 73.2$\pm$1.6 & 69.7   \\
				SpeGCL (Ours)                        & \textbf{72.3$\pm$1.5} & \textbf{77.7$\pm$0.7} & \textbf{65.6$\pm$0.7} & \textbf{63.9$\pm$1.5} & \underline{76.6$\pm$2.3} & \textbf{78.4$\pm$1.1} & 75.9$\pm$0.4 & \textbf{72.8$\pm$0.6} & 76.7$\pm$1.4 & \textbf{73.3}    \\ \bottomrule
		\end{tabular}}
	\end{table*}
	
	\subsection{Experimental Details}
	\label{sec:experimental}
	
	In our graph classification experiments, we adopted FourierGCN as the encoder and selected the number of layers from $\{4, 8, 12\}$ and the hidden dimensions from $\{32, 512\}$. For the optimizer, we used Adam \cite{kingma2014adam} and selected the learning rate from $\{10^{-3}, 10^{-4}, 10^{-5}\}$, and selected the epoch number from $\{60, 100\}$. Following previous work \cite{sun2019infograph}, we fed the generated graph embeddings as input to the SVM classifier to evaluate the performance of the graph embeddings in downstream classification tasks. To ensure the generalization and robustness of the model on different datasets, we used the cross-validation method to independently adjust the parameters of the classifier. Cross-validation divides the dataset into multiple subsets, uses one part for validation each time, and uses the rest for training to repeatedly evaluate the performance of the model. Cross-validation can effectively avoid overfitting and help select the best hyperparameter combination, thereby improving the adaptability and classification effect of the classifier on different samples. We performed all experiments on a high-performance device equipped with a 24GB NVIDIA GeForce RTX 4090 graphics card.
	
	\textbf{Unsupervised Learning.} Table \ref{tab:unsuperise} shows the comparison of the unsupervised graph learning classification effect of our proposed method on TUDataset with other advanced methods. The experimental results show that our model has achieved impressive graph classification results on multiple datasets, especially on PROTEINS, NCI1, IMDB-binary and REDDIT-Multi-5K datasets, where we have achieved the best classification accuracy. In addition, on MUTAG, DD and REDDIT binary datasets, our method also performs well and achieves suboptimal results, surpassing most existing comparative learning methods, including GraphCL, JOAO and AD-GCL. The performance improvement can be attributed to the fact that our proposed SpeGCL model can effectively capture the high-frequency information in the graph structure, thereby optimizing the representation after graph data augmentation. The experimental results show that our method is widely applicable and competitive on different types of datasets, further proving the key role of high-frequency information in GCLs.

	\begin{table*}[htbp]
		\centering
		\caption{Overall comparison with existing semi-supervised learning methods on multiple graph classification datasets. We report accuracy results as mean $\pm$ std.}
		\label{tab:unsupervise}
		\renewcommand\arraystretch{1}
		\setlength{\tabcolsep}{2mm}{
			\begin{tabular}{@{}lcccccccc@{}}
				\toprule
				\textbf{Model}        & PROTEINS   & DD         & NCI1       & COLLAB     & GITHUB     & IMDB-B     & REDDIT-B   & REDDIT-M-5K \\ \midrule
				\textit{Full Data}    & 79.56±1.43 & 81.98±2.84 & 84.98±1.18 & 84.59±0.83 & 67.83±1.37 & 78.58±3.23 & 89.58±1.93 & 56.47±1.17  \\ \midrule
				10\% Data             & 69.72±6.71 & 74.36±5.86 & 75.16±2.07 & 74.34±2.00 & 61.05±1.57 & 64.80±4.92 & 76.75±5.60 & 49.71±3.20  \\
				10\% GCA \cite{zhu2021graph}             & 73.85±5.56 & 76.74±4.09 & 68.73±2.36 & 74.32±2.30 & 59.24±3.21 & 73.70±4.88 & 77.15±6.96 & 32.95±10.89 \\
				10\% GraphCL Aug Only \cite{you2020graph}  & 70.71±5.63 & 76.48±4.12 & 70.97±2.08 & 73.56±2.52 & 59.80±1.94 & \underline{71.10±5.11} & 76.45±4.83 & 47.33±4.02  \\
				10\% GraphCL \cite{you2020graph}     & 74.21±4.50 & 76.65±5.12 & 73.16±2.90 & 75.50±2.15 & 63.51±1.02 & 68.10±5.15 & 78.05±2.65 & 48.09±1.74  \\
				10\% JOAOv2  \cite{you2021graph}         & 73.31±0.48 & 75.81±0.73 & \underline{74.86±0.39} & 75.53±0.18 & \textbf{66.66±0.60} & -          & 88.79±0.65 & 52.71±0.28  \\
				10\% AD-GCL  \cite{suresh2021adversarial}         & 73.96±0.47 & \underline{77.91±0.73} & \textbf{75.18±0.31} & 75.82±0.26 & -          & -          & \textbf{90.10±0.15} & 53.49±0.28  \\ 
				10\% AutoGCL \cite{yin2022autogcl} & \underline{75.65±2.40} & 77.50±4.41 & 73.75±2.25 & \underline{77.16±1.48} & 62.46±1.51 & \textbf{71.90±4.79} & 79.80±3.47 & 49.91±2.70 \\ 
				10\% SEGA \cite{wu2023sega} & 74.65±0.54 & 76.33±0.43 &  75.09±0.22 & 75.18±0.22 &  66.01±0.66 & - & 89.40±0.23 &  \underline{53.73±0.28} \\ 
				10\% SpeGCL (Ours) & \textbf{77.77±1.79} & \textbf{79.47±3.01} & 74.28±1.74 & \textbf{80.27±1.59} & \underline{66.31±0.84} & 69.59±5.27 & \underline{89.96±2.51} & \textbf{56.15±3.75} \\ 
				\bottomrule
		\end{tabular}}
	\end{table*}
	
	\textbf{Transfer Learning.} Table \ref{tab:transfer} shows the detailed comparison results of different methods on the MoleculeNet dataset in the transfer learning experimental environment. The experimental results show that SpeGCL method has achieved significant performance improvements on most datasets (e.g., BBBP, ClinTox, MUV and BACE), showing its excellent performance in molecular graph representation learning tasks. In contrast, the existing state-of-the-art model AD-GCL failed to achieve comparable results with SpeGCL on multiple datasets. In particular, the advantages of SpeGCL are more obvious in tasks with large noise or complex structures. The performance improvement may be due to the unique design of SpeGCL, which makes it more robust and adaptable when facing dynamic changes in node features. In addition, our proposed SpeGCL method no longer needs to rely on positive samples to guide model learning like traditional methods. On the contrary, as long as there are enough negative samples, the model can also achieve effective convergence. Our method breaks the inherent assumption of positive and negative sample balance in traditional supervised learning and provides a new idea for the design of contrast loss in GCLs.
	
	As shown in the Table \ref{tab:tran}, we conducted transfer learning experiments on more datasets (i.e., PPI-306K and ZINC 2M). Experimental results also show that our method can achieve optimal results on most data sets.
	
	\textbf{Semi-Supervised Learning.} In the semi-supervised experimental setting, as shown in Table \ref{tab:unsupervise}, we tested the semi-supervised tasks with a label rate of 10\%. The experimental results show that the proposed SpeGCL method outperforms the previous baseline method in most cases, or performs comparable to the existing state-of-the-art methods (SOTA). The performance improvement of SpeGCL may be mainly attributed to its excellent ability to fully utilize the node label information. In the semi-supervised setting, due to the limited amount of labeled data, how to effectively use a small amount of label information to improve the feature representation ability of unlabeled nodes is a key challenge. SpeGCL can better integrate a small amount of label information through its clever design in Fourier space, so that it has a positive impact on the feature representation of the entire graph. SpeGCL not only enhances the representation learning effect of labeled nodes, but also indirectly improves the feature learning ability of unlabeled nodes, enabling the model to maintain high accuracy and robustness with less supervised information. In addition, SpeGCL effectively mines the local and global information in the graph in Fourier space, enabling the model to obtain better feature embedding during the model learning process.
	
	\begin{table*}[htbp]
		\centering
		\caption{Ablation study with existing semi-supervised learning methods on multiple graph classification datasets. We report accuracy results as mean $\pm$ std.}
		\label{tab:ablation}
		\renewcommand\arraystretch{1}
		\setlength{\tabcolsep}{1.5mm}{
			\begin{tabular}{@{}lcccccccc@{}}
				\toprule
				Model & MUTAG     & \multicolumn{1}{c}{PROTEINS} & \multicolumn{1}{c}{DD} & \multicolumn{1}{c}{NCI1} & COLLAB & IMDB-B & REDDIT-B & REDDIT-M-5K \\ \midrule
				w Pos/Neg &  \underline{89.76$\pm$1.18}                           &   72.58$\pm$0.87                     &       79.05$\pm$0.43                   &   \textbf{82.86$\pm$1.34}     &   \textbf{71.48$\pm$0.42}     &   \underline{75.76$\pm$0.58}       &  \textbf{92.31$\pm$0.21}    &   58.99$\pm$0.37    \\
				w/o Neg   &  88.75$\pm$1.47                            &     71.79$\pm$1.27                   &    \underline{80.41$\pm$0.71}                      &   82.02$\pm$0.85     &    69.11$\pm$0.26    &  75.48$\pm$1.55        &  89.34$\pm$0.57    & \underline{59.17$\pm$0.21}      \\
				w/o Pos   & \textbf{90.86±2.74} & \textbf{75.05±1.23} & \textbf{81.23±0.94} & \underline{82.14±1.12} & 70.00±0.38
				& \textbf{76.57±1.95} & \underline{91.71±0.31} & \textbf{59.44±0.18}             \\ 
				w/o FourierGNN &  87.97±1.85 & \underline{73.44±0.97} &  78.49$\pm$0.68 & 77.14$\pm$ 1.08 & \underline{71.29$\pm$0.58} &  72.38±0.83  &   82.05±0.89  & 56.82$\pm$0.33   \\ \bottomrule
		\end{tabular}}
	\end{table*}
	
	\begin{figure*}[htbp]
		\centering
		\includegraphics[width=1\linewidth]{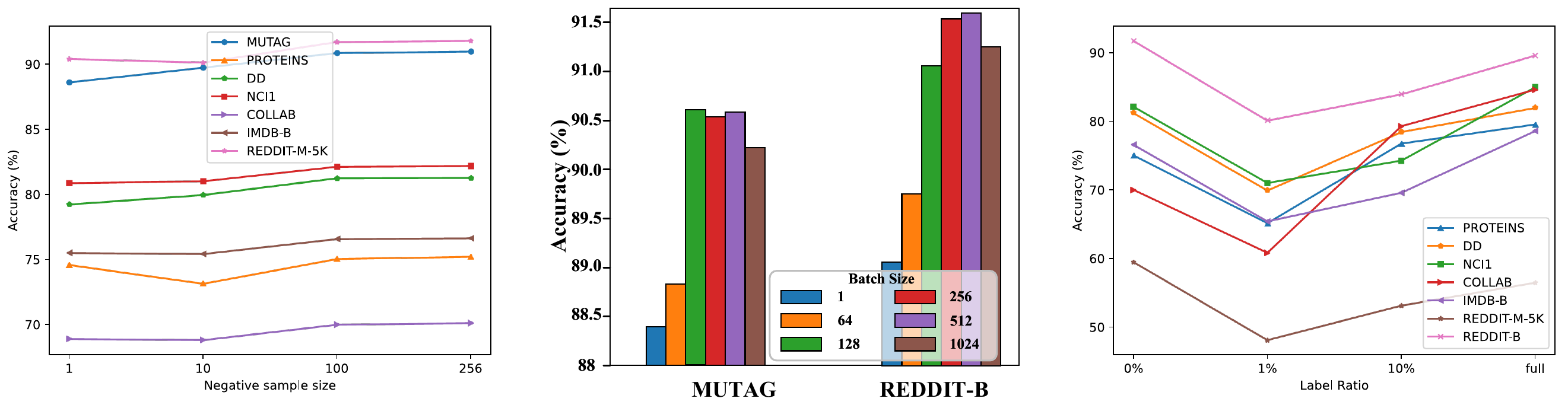}
		\caption{\textbf{Left:} Exploring the impact of the number of negative samples on graph classification performance under unsupervised experimental conditions. \textbf{Middle:} Explore the impact of the number of batch sizes on graph classification performance under unsupervised experimental conditions. \textbf{Right:} Exploring the impact of different label ratios on graph classification performance.}
		\label{fig:hyper}
		\vspace{-4mm}
	\end{figure*}
	
	\textbf{Ablation Study.} Current research generally believes that positive samples in GCL are crucial and indispensable for model training. However, we unexpectedly found that the GCL method can achieve satisfactory performance even without any positive sample pairs. To verify this observation, we performed a series of ablation experiments. As shown in Table \ref{tab:ablation}, in most graph classification datasets, the accuracy gap between using positive and negative sample pairs and not using any positive samples (NO Pos) is relatively small. Sometimes, even in the absence of positive samples, the accuracy of graph classification can surpass that of using pairs of positive and negative samples. This finding suggests that the role of positive samples in GCL may be overestimated. Further analysis of the experimental results shows that in GCL, removing positive samples has minimal impact on the performance of downstream benchmark tests, which demonstrates GCL's ability to utilize negative samples and the model's adaptive ability in the absence of positive samples. The emergence of this phenomenon may be due to the GCL model's learning ability and sensitivity to negative samples. As a result, even if there is a lack of positive samples, the model can still obtain enough information from negative samples to complete the graph classification task. This discovery is of great significance for improving graph comparison learning algorithms and understanding the internal working mechanism of the model and also provides new inspiration for future research directions in GCLs. 
	
	In addition, we also analyze the experimental results without using FuorierGNN. Specifically, we replace FuorierGNN with GIN as the model’s encoder, and only use negative samples to build the contrastive loss. The experimental results are shown in Table \ref{tab:ablation}. The accuracy of graph classification using GIN is significantly higher than that of FourierGNN. The experimental results show that high-frequency information promotes model learning.

	\textbf{Impact of hyper-parameters.} The main hyper-parameters in SpeGCL are negative sample size and batch size (affecting the capacity of negative samples). As pointed out by Theorem 1 and Theorem 2, the error term of contrastive loss decays with O($M^{-\frac{2}{3}}$), which shows the importance of expanding the number of negative samples. Therefore, we explored the impact of different batch sizes ($\{1, 64,128,256,512,1024\}$) and the number of negative samples ($\{1, 10, 100, 256\}$) on graph classification accuracy. The experimental results are shown in Figure \ref{fig:hyper}. Fix batch size to 128. When the number of negative samples is between 1 and 10, the performance improvement is not obvious. But when the number of negative samples grows to 100, the improvement in graph classification accuracy becomes significant. When the number of negative samples is increased to 256, the performance of the model does not improve significantly. Fixing the number of negative samples to 100, the accuracy of graph classification improves steadily as the batch size range increases from 1 to 512. But when the batch size is 1024, the performance of the model decreases. Furthermore, we also explored the impact of different label ratios on graph classification performance. Experimental results show that unsupervised learning is better than semi-supervised learning, and under semi-supervised learning conditions, the performance of the model increases as the proportion of labels increases.
	
	\begin{table}
		\centering
		\caption{Efficiency comparison on PROTEINS and COLLAB graph datasets. In particular, both GraphCL and our method use masking nodes and edge perturbations for data augmentation. We report the model's running time and memory.}
		\label{tab:runningtime}
		\renewcommand\arraystretch{1}
		\setlength{\tabcolsep}{5mm}{
			\begin{tabular}{lccc}
				\toprule
				Dataset                   & Algorithm    & Training & Memory    \\ \midrule
				\multirow{3}{*}{PROTEINS} & GraphCL      & 111s      & 1231M  \\
				& JOAOv2       & 4088S   & 1403M  \\
				& SpeGCL & \textbf{46s}      & \textbf{1175M}   \\ \midrule
				\multirow{3}{*}{COLLAB}   & GraphCL      & 1033s   & 10199M \\
				& JOAOv2       & 10742s  & 7303M \\
				& SpeGCL & \textbf{378s}     & \textbf{6547M}    \\ \bottomrule
		\end{tabular}}
		\vspace{-3mm}
	\end{table}

	\textbf{Memory and Computation Efficiency.} In Table \ref{tab:runningtime}, we compare the performance of the proposed SpeGCL method with two baseline methods, GraphCL and JOAOv2, in terms of training time and memory overhead. Specifically, training time refers to the total time it takes for the model to complete all training steps in the training phase in a semi-supervised experimental setting. This includes forward propagation, backward propagation, and gradient updates. In addition to training time, we also analyze the memory overhead of each model. Memory overhead mainly refers to the total memory resources occupied by model parameters and all hidden layer representations of batch data during training. Specifically, SpeGCL is more efficient in training time compared to other methods, such as GraphCL and JOAOv2. Secondly, SpeGCL also shows advantages in memory overhead. Due to its more efficient graph augmentation and feature learning mechanism, SpeGCL uses more streamlined model parameters while maintaining high performance, and effectively compresses the dimensions of hidden representations. Therefore, SpeGCL occupies less video memory and memory resources than GraphCL and JOAOv2 under the same batch size and model structure.

	\section{Conclusions}
	\label{sec:con}
	
	In this paper, we explore the application of Fourier graph networks for graph classification from the perspective of graph spectrum. To solve the problem that existing methods cannot fully utilize the high-frequency information of node features and require time-consuming construction of positive and negative sample pairs, we propose a novel spectral graph contrastive learning framework without positive samples (SpeGCL). Specifically, SpeGCL uses Fourier operations to obtain high-frequency and low-frequency information of node features. While the graph view performs contrastive learning to retain the most expressive local context information in the nodes. Furthermore, SpeGCL uses only negative samples to optimize the embedding representation of the graph. We also theoretically demonstrate the rationality of using only negative samples on GCL. Extensive experiments have been conducted to prove the superiority of our SpeGCL framework over the state-of-the-art GCLs.
	

	\bibliographystyle{IEEEtran}
	\bibliography{example_paper}

	%
	
	
	\section*{Appendix}
	
	\subsection{Convolution Theorem}
	\label{sec:conv}
	
	The convolution theorem \cite{katznelson2004introduction} is a core concept in the field of Fourier transforms, which reveals the direct connection between the convolution operation in the time domain and the product operation in the frequency domain. Specifically, the convolution theorem states that if there are two signals, such as an input signal $x[n]$ and an impulse response $h[n]$ of a system or filter, their convolution result y[n] in the time domain can be obtained by performing a point-by-point product of the Fourier transforms of the two signals and then performing an inverse Fourier transform on the result:
	\begin{equation}
		\mathcal{F}\{(x \ast h)[n]\} = \mathcal{F}\{x[n]\} \cdot \mathcal{F}\{h[n]\}
	\end{equation}
	where $\mathcal{F}\{\cdot\}$ represents the Fourier transform, $(x \ast h)[n]$ is the convolution of a signal $x[n]$ and a filter $h[n]$.

	\subsection{Proof of Theorem 1.}
	\label{sec:proof1}
	
	\textbf{Theorem 1.} For a fixed $\tau > 0$ , when the number of negative samples $M \rightarrow \infty$, the contrastive loss $\mathcal{L}_{NCE}$ converges and the absolute deviation decays with $O(M^{-2/3}$). If there exists a perfectly uniform encoder $f$, it is able to obtain the minimum value.
	
	\textit{Proof.} Note that for any $x,y\in\mathbb{R}^n\text{ and }\{x_i^-\}_{i=1}^M\stackrel{\text{i.i.d.}}{\sim}p_{\text{data}}$, according to the strong law of large number (SLLN) and the continuous mapping theorem we have 
	\begin{equation}
		\begin{aligned}
			&\lim_{M\to\infty}\log\left(\frac{1}{M}e^{f(x)^{\mathsf{T}}f(y)/\tau}+\frac{1}{M}\sum_{i=1}^{M}e^{f(x_{i}^{-})^{\mathsf{T}}f(x)/\tau}\right) \\
			&=\log\underset{x^{-}\sim p_{\mathsf{data}}}{\operatorname*{\mathbb{E}}}\left[e^{f(x^{-})^{\mathsf{T}}f(x)/\tau}\right]
		\end{aligned}
	\end{equation}
	
	According to the Dominated Convergence Theorem (DCT) \cite{blackwell1963converse}, we can derive
	
	\begin{equation}
		\begin{aligned}
			&\left|\left(\lim_{M\to\infty}\mathcal{L}{ ( f ; \tau , M ) - \log M }\right)-\left(\mathcal{L}{ ( f ; \tau , M ) - \log M }\right)\right| \\
			&=\left|\begin{matrix}\mathbb{E}\\(x,y)\underset{i=1}{\operatorname*{\sim}}p_{\mathsf{pos}}\\\{x_i^-\}_{i=1}^M\overset{\mathrm{i.i.d.}}{\operatorname*{\sim}}p_{\mathsf{data}}\end{matrix}\left[\log\mathbb{E}_{x^-\thicksim p_{\mathsf{data}}}\left[e^{f(x^-)^\intercal f(x)/\tau}\right]\right.\right. \\&\left.\left.-\log\left(\frac{1}{M}e^{f(x)^{\mathsf{T}}f(y)/\tau}+\frac{1}{M}\sum_{i=1}^{M}e^{f(x_i^{-})^{\mathsf{T}}f(x)/\tau}\right) \right]
			\right| \\
			&\leq e^{1/\tau}\begin{matrix}\mathbb{E}\\(x,y)\underset{i=1}{\operatorname*{\sim}}p_{\mathsf{pos}}\\\{x_i^-\}_{i=1}^M\overset{\mathrm{i.i.d.}}{\operatorname*{\sim}}p_{\mathsf{data}}\end{matrix}\left[\left|\mathbb{E}_{x^-\thicksim p_{\mathsf{data}}}\left[e^{f(x^-)^\intercal f(x)/\tau}\right]\right.\right. \\ &\left.\left.-\left(\frac{1}{M}e^{f(x)^{\mathsf{T}}f(y)/\tau}+\frac{1}{M}\sum_{i=1}^{M}e^{f(x_i^{-})^{\mathsf{T}}f(x)/\tau}\right) \right|\right]
			\\
			&\le\frac{1}{M}e^{2/\tau}+e^{1/\tau} \underset{x,\{x_i^-\}_{i=1}^M\overset{\mathrm{i.i.d.}}{\operatorname*{\sim}}p_\text{data}}{\mathbb{E}} \left[\left|\mathbb{E}_{x^-\thicksim p_{\mathsf{data}}}\left[e^{f(x^-)^\intercal f(x)/\tau}\right]\right.\right. \\ & \left.\left.-\frac{1}{M}\sum_{i=1}^{M}e^{f(x_i^{-})^{\mathsf{T}}f(x)/\tau} \right|\right]  =\frac{1}{M}e^{2/\tau}+\mathcal{O}(M^{-2/3})
		\end{aligned}
	\end{equation}
	where the first inequality is obtained based on the intermediate value theorem and the absolute derivative of the logarithm of the upper bound $e^{1/\tau}$ between two points, and the last equation is obtained based on the Berry-Esseen theorem and considering the bounded supportability of $e^{f(x^-_i)^Tf(x)/\tau}$. Specifically, for an independent and identically distributed random variable $Y_i$ with bounded support $\subset [-a,a]$, zero mean and variance $\sigma^2_Y \leq a^2$, we have:
	
	\begin{gather}
		\begin{aligned}
			&\mathbb{E}\left[\left|\frac1M\sum_{i=1}^MY_i\right|\right]
			=\frac{\sigma_Y}{\sqrt{M}}\mathbb{E}\left[\left|\frac1{\sqrt{M}\sigma_Y}\sum_{i=1}^MY_i\right|\right]  \\
			&=\frac{\sigma_Y}{\sqrt M}\int_0^{\frac{a\sqrt M}{\sigma_Y}}\mathbb{P}\left[\left|\frac1{\sqrt M\sigma_Y}\sum_{i=1}^MY_i\right|>x\right]\mathrm{d}x \\
			&\leq\frac{\sigma_Y}{\sqrt{M}}\int_0^{\frac{a\sqrt{M}}{\sigma_Y}}\mathbb{P}\left[|\mathcal{N}(0,1)|>x\right]+\frac{C_a}{\sqrt{M}}\operatorname{d}x& (\text{Berry-Esseen})  \\
			&\leq\frac{\sigma_Y}{\sqrt{M}}\left(\frac{aC_a}{\sigma_Y}+\int_0^\infty\mathbb{P}\left[|\mathcal{N}(0,1)|>x\right]\mathrm{d}x\right) \\
			&=\frac{\sigma_Y}{\sqrt{M}}\left(\frac{aC_a}{\sigma_Y}+\mathbb{E}\left[|\mathcal{N}(0,1)|\right]\right) \\
			&\leq\frac{C_a}{\sqrt{M}}+\frac a{\sqrt{M}}\mathbb{E}\left[|\mathcal{N}(0,1)|\right]=\mathcal{O}(M^{-2/3})
	\end{aligned}\end{gather}
	
	\subsection{Proof of Proposition 1}
	\textbf{Proposition 1.} For fixed $\tau > 0$ , the upper limit of $\mathcal{L}_U$ is always controlled by $\mathcal{L}_{NCE}$.
	
	\textit{Proof.} We assume that nodes are independently and identically distributed, and then we sample negative samples from the augmented graph to get:
	
	\begin{gather}\begin{aligned}
			\mathcal{L}_U& =\underset{\{y_i^-\}_{\boldsymbol{i}=1}^M\overset{\mathrm{i.i.d.}}{\sim}\boldsymbol{p_y}}{\mathbb{E}}[-\log\frac{e^{\frac1\tau}}{e^{\frac1\tau}+\sum_ie^{f(x)^{\mathsf{T}}f(y_i^-)/\tau}}]  \\
			&\leq \underset{\substack{(x, y) \sim p_{\text {pos }} \\ \{{y}_{{i}}^-\}_{\boldsymbol{i}=1}^{\boldsymbol{M}}\sim\boldsymbol{p}_{{y}}}}{\mathbb{E}}\left[-\log\frac{ef(x)^\mathsf{T}f(y)/\tau}{e^{f(x)^\mathsf{T}f(y)/\tau}+\sum_ie^{f(x)^\mathsf{T}f(y_i^-)/\tau}}\right]\\&=\mathcal{L}_{\mathrm{NCE}}. \\
	\end{aligned}\end{gather}
	
	On the other hand,
	\begin{equation}
		\begin{aligned}
			& \mathcal{L}_{\mathrm{NCE}} \\ & \leq \underset{\substack{(x, y) \sim p_{\text {pos }} \\
					\left\{y_i^{-}\right\}_{i=1}^{M }\overset{\text { i.i.d. }}{\sim} p_{\mathrm{y}}}}{\mathbb{E}}\left[-\log \left(\frac{e^{\min \left(f(x)^{\top} f(y)\right) / \tau}}{e^{\min \frac{f(x)^{\top} f(y)}{\tau}}+\sum_i e^{\frac{f(x)^{\top} f(y_i^{-})}{ \tau}}}\right)\right] \\
			& \leq \underset{\substack{(x, y) \sim p_{\text {pos }} \\
					\left\{y_i^{-}\right\}_{i=1}^M \text { i.i.d. }}}{\mathbb{E}}\left[-\log \left(\frac{e^{\min \left(f(x)^{\top} f(y)\right) / \tau}}{e^{\frac{1}{\tau}+\sum_i e^{f(x)^{\top} f\left(y_i^{-}\right) / \tau}}}\right)\right] \\
			& \leq \mathcal{L}_{{U}}+\frac{1}{\tau}\left[1-\min _{(x, y) \sim p_{\text {pos }}}\left(f(x)^{\top} f(y)\right)\right] \\
			&
		\end{aligned}
	\end{equation}
	
	\subsection{Proof of Theorem 2}
	\label{sec:proof2}
	
	\textbf{Theorem 2.} For the given constant $\tau\in\mathbb{R}^+$, $L_U$ still converges to the same limit as NCE loss, and the absolute deviation decays by $O(M^{-2/3})$.
	
	\textit{Proof.} We follow the the outline of Wang’s proof \cite{wang2020understanding}. According to the last equality is by the strong law of large numbers (SLLN), and the continuous mapping theorem we have:
	
	\begin{gather}
		\begin{aligned}
			& \operatorname*{lim}_{M\to\infty}\mathcal{L}(f;\tau,M)-\operatorname{log}M  \\
			&=\underset{(x,y)\sim p_{\mathbf{pss}}}{\operatorname*{\mathbb{E}}}\left[-f(x)^\mathsf{T}f(y)/\tau\right]\\&+\underset{\begin{array}{c}\{x_i^-\}_{i=1}^M\overset{\mathrm{i.i.d}}{\operatorname*{\sim}}p_{\mathbf{data}}\\\end{array}}{\operatorname*{\mathbb{E}}}\left[\log\left(\frac1Me^{f(x)^\mathsf{T}f(y)/\tau}\right.\right.\\&\left.\left.+\frac1M\sum_{i=1}^Me^{f(x_i^-)^\mathsf{T}f(x)/\tau}\right)\right] \\
			&=\mathbb{E}_{(x,y)\sim p_{\mathsf{pos}}}[-f(x)^\mathsf{T}f(y)/\tau]\\&+\mathbb{E}\left[\lim_{M\to\infty}\log\left(\frac1Me^{f(x)^\mathsf{T}f(y)/\tau}+\frac1M\sum_{i=1}^Me^{f(x_i^-)^\mathsf{T}f(x)/\tau}\right)\right] \\
			&                                                                =-\frac1{\tau}\underset{(x,y)\sim p_{\mathsf{pos}}}{\operatorname*{\mathbb{E}}}\left[f(x)^\mathsf{T}f(y)\right]\\&+\underset{x\sim p_{\mathsf{data}}}{\operatorname*{\mathbb{E}}}\left[\underset{x^-\sim p_{\mathsf{data}}}{\operatorname*{\log}}{\operatorname*{\mathbb{E}}}\left[e^{f(x^-)^\mathsf{T}f(x)/\tau}\right]\right]
		\end{aligned}
	\end{gather}
	
	Therefore,
	\begin{equation}
		\begin{aligned}
			& \lim _{M \rightarrow \infty}\left[\mathcal{L}_{{U} \mid \mathrm{x}}\left(f ; \tau, M, p_{\mathrm{y}}\right)-\log M\right] \\
			& =-\frac{1}{\tau} \underset{(x, y) \sim p_{\text {pos }}}{\mathbb{E}}\left[f(x)^{\top} f(y)\right] \\
			& +\lim _{M \rightarrow \infty} \underset{(x, y) \sim p_{\text {pos }}}{\mathbb{E}}\left[\log \left(\frac{1}{M} e^{\frac{f(x)^{\top} f(y)}{ \tau}}+\frac{1}{M} \sum_i e^{\frac{f(x)^{\top} f\left(y_i^{-}\right)}{ \tau}}\right)\right] \\
			& =-\frac{1}{\tau} \underset{(x, y) \sim p_{\text {pos }}}{\mathbb{E}}\left[f(x)^{\top} f(y)\right]\\&+\underset{x \stackrel{\text { i.i.d. }}{\sim} \text { px }}{\mathbb{E}}\left[\log \underset{y^{- \text {i.i.d. }} p_y}{\mathbb{E}}\left[e^{f(x)^{\top} f\left(y^{-}\right) / \tau}\right]\right] \\
			&
		\end{aligned}
	\end{equation}
	
	The convergence speed is derived as follows:
	
	On the one hand:
	\begin{equation}\begin{aligned}
			&\mathcal{L}_{U|x}(f;\tau,M,p_{Y})-\log M-\operatorname*{lim}_{M\to\infty}[\mathcal{L}_{U|x}(f;\tau,M,p_{Y})-\log M] \\
			&\leq \underset{\substack{x \overset{\mathrm{i.i.d.}} \sim p_x \\ \{{y_i^-}\}_{{i=}1}^{\boldsymbol{M}}\overset{\mathrm{i.i.d.}}{\operatorname*{\sim}}\boldsymbol{p_y} }}{\mathbb{E}}\left[\log\left(\frac 1 Me^{1/\tau}+\frac1M\sum_ie^{f(x)^{\mathsf{T}}f(y_i^-)/\tau}\right)\right] \\
			& -\underset{x \stackrel{\text { i.i.d. }}{\sim}p_x}{\mathbb{E}}\left[\log \underset{y^{- \overset{\text {i.i.d. }}{\operatorname*{\sim}}} p_y}{\mathbb{E}}\left[e^{f(x)^{\top} f\left(y^{-}\right) / \tau}\right]\right] \\
			& \leq \underset{x \stackrel{\text { i.i.d. }}{\sim}p_x}{{\mathbb{E}}}\left[\log \underset{y^{-} \underset{\sim}{- \text { i.i.d. }} p_y}{\mathbb{E}}\left[\left(\frac{1}{M} e^{1 / \tau}+e^{f(x)^{\top} f\left(y^{-}\right) / \tau}\right)\right]\right.
			\\& \left.-\log \underset{y^{- \overset{\text {i.i.d. }}{\operatorname*{\sim}}} p_y}{\mathbb{E}}\left[e^{f(x)^{\top} f\left(y^{-}\right) / \tau}\right]\right] \\
			&\leq\mathbb{E}_{\begin{array}{c}x \stackrel{\text { i.i.d. }}{\sim}p_x\\\end{array}}\left[\frac 1 Me^{2/\tau}\right]=\frac 1 Me^{2/\tau}
	\end{aligned}\end{equation}
	
	On the other hand:
	\begin{equation}
		\begin{aligned}
			& \lim _{M \rightarrow \infty}\left[\mathcal{L}_{U \mid \mathrm{x}}\left(f ; \tau, M, p_{\mathrm{y}}\right)-\log M\right] \\
			& \leq e^{1 / \tau} \underset{(x, y) \sim p_{\text {pos }}}{\mathbb{E}}\left[\left|\underset{y^{- \overset{\text {i.i.d. }}{\operatorname*{\sim}}} p_y}{{\mathbb{E}}}\left[e^{f(x)^{\top} f\left(y^{-}\right) / \tau}\right]\right.\right.\\&\left.\left.-\left(\frac{1}{M} e^{f(x)^{\top} f(y) / \tau}+\frac{1}{M} \sum_i e^{f(x)^{\top} f\left(y_i^{-}\right) / \tau}\right)\right|\right] \\
			&\leq\frac{1}{M}e^{2/\tau}+e^{1/\tau} \underset{\{y _ {i}^{\boldsymbol{-}}\}_{\boldsymbol{i}=1}^{\boldsymbol{M}}\overset{\mathrm{i.i.d.}}{\operatorname*{\sim}}{p}_{{y}}}{\underset{(x, y) \sim p_{\text {pos }}}{\mathbb{E}}} \left[\underset{\boldsymbol{y^{\overset{-\mathrm{i.i.d.}}{\operatorname*{\sim}}{P}{y}}}}{\mathbb{E}}| \left[e^{f(x)^{\mathsf{T}}f(y^{-})/\tau}\right]\right.\\&\left.-\frac1M\sum_ie^{f(x)^{\mathsf{T}}f(y_i^{-})/\tau} \right] \\
			& \leq\frac 1  Me^{2/\tau}+\frac54M^{-\frac23}e^{\frac1\tau}\left(e^{\frac1\tau}-e^{-\frac1\tau}\right)
		\end{aligned}
	\end{equation}
	
	Therefore, $L_U$ still converges to the same limit as NCE loss, and the absolute deviation decays by $O(M^{-2/3})$.

	\vfill
	
\end{document}